\documentclass[10pt,twocolumn,letterpaper]{article}

\usepackage{iccv}
\usepackage{times}
\usepackage{epsfig}

\usepackage{graphicx}
\usepackage{amsmath}
\usepackage{amssymb}
\usepackage{booktabs}
\usepackage{amsthm}
\usepackage{subcaption}
\usepackage{bm}
\usepackage{paralist}
\usepackage[numbers,compress]{natbib}

\usepackage{algorithm2e}
\RestyleAlgo{ruled}

\theoremstyle{definition}

\usepackage{extarrows}
\newcommand{\eqdef}{\xlongequal{\text{def}}}%

\newcommand{\mypara}[1]{\vspace{2pt}\noindent\textbf{#1}}

\DeclareMathOperator{\pos}{pos}
\DeclareMathOperator{\blind}{blind}
\DeclareMathOperator{\intersect}{intersect}
\DeclareMathOperator{\slack}{slack}
\DeclareMathOperator{\ddiff}{d\_diff}
\DeclareMathOperator{\view}{view}
\DeclareMathOperator{\clamp}{clamp}
\DeclareMathOperator{\comp}{comp}
\DeclareMathOperator{\risk}{risk}
\DeclareMathOperator{\MSE}{MSE}
\DeclareMathOperator{\Attention}{Attention}
\DeclareMathOperator{\Softmax}{Softmax}

\usepackage[breaklinks=true,bookmarks=false]{hyperref}

\iccvfinalcopy 

\def\iccvPaperID{XXX} 

\ificcvfinal\fi

\begin{document}

\title{Exploiting Proximity-Aware Tasks for Embodied Social Navigation}

\author{Enrico Cancelli$^{1}$\thanks{Both authors contributed equally to this work.}
\quad
Tommaso Campari$^{1,2}$\footnotemark[1]
\quad
Luciano Serafini$^{2}$
\quad
Angel X. Chang$^{3}$
\quad
Lamberto Ballan$^1$\\\\
$^{1}$ University of Padova \quad $^{2}$ Fondazione Bruno Kessler (FBK) \quad $^{3}$ Simon Fraser University
}


\maketitle
\ificcvfinal\thispagestyle{empty}\fi

\begin{abstract}
Learning how to navigate among humans in an occluded and spatially constrained indoor environment, is a key ability required to embodied agent to be integrated into our society. In this paper, we propose an end-to-end architecture that exploits Proximity-Aware Tasks (referred as to Risk and Proximity Compass) to inject into a reinforcement learning navigation policy the ability to infer common-sense social behaviors. To this end, our tasks exploit the notion of immediate and future dangers of collision. Furthermore, we propose an evaluation protocol specifically designed for the Social Navigation Task in simulated environments. This is done to capture fine-grained features and characteristics of the policy by analyzing the minimal unit of human-robot spatial interaction, called Encounter. We validate our approach on Gibson4+ and Habitat-Matterport3D datasets.
\vspace{-10pt}
\end{abstract}


\section{Introduction}
\label{sec:intro}
\begin{figure}[t]
    \centering
    \includegraphics[width=.49\linewidth]{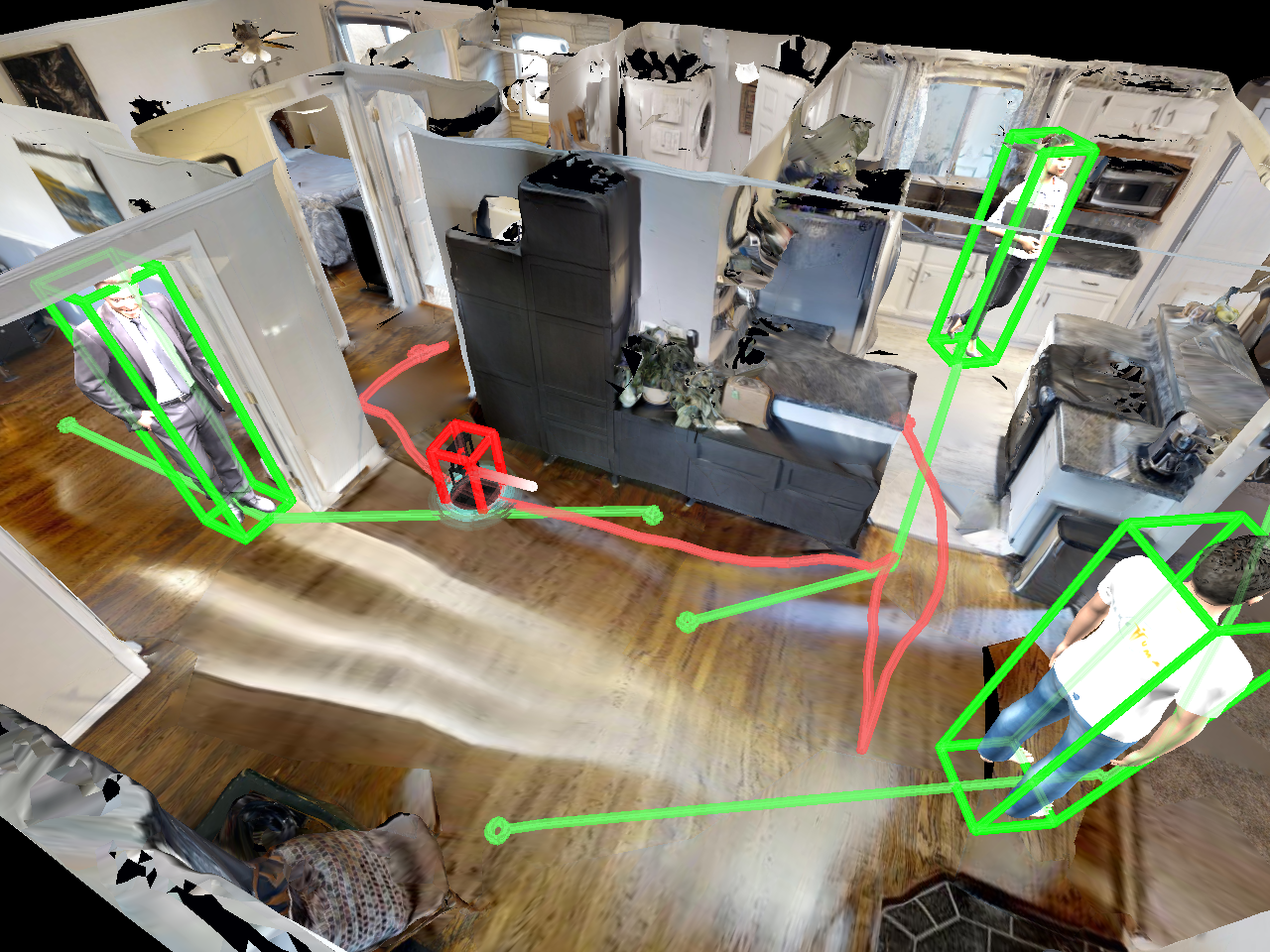}
    \includegraphics[width=.49\linewidth]{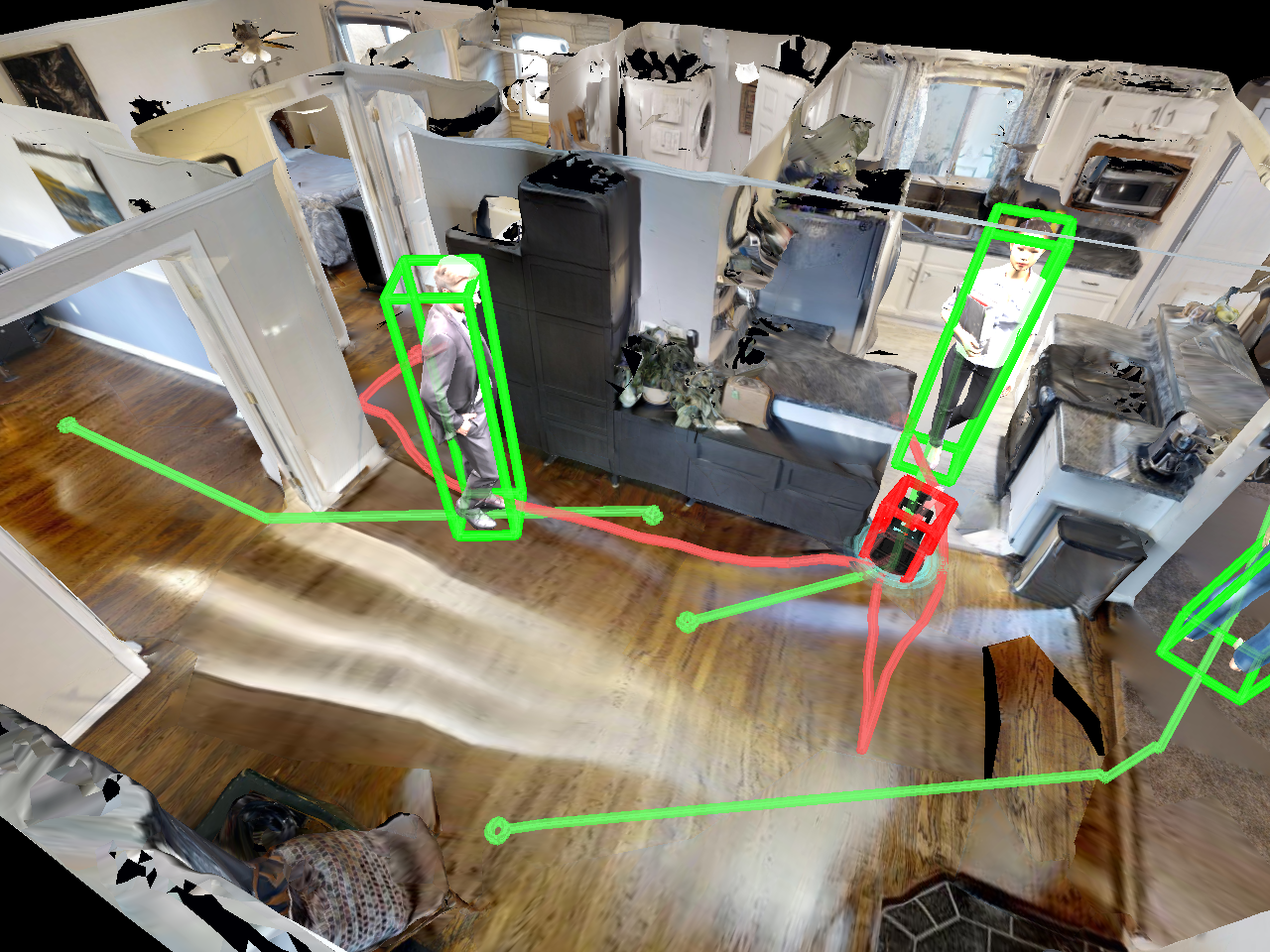}
    \includegraphics[width=.49\linewidth]{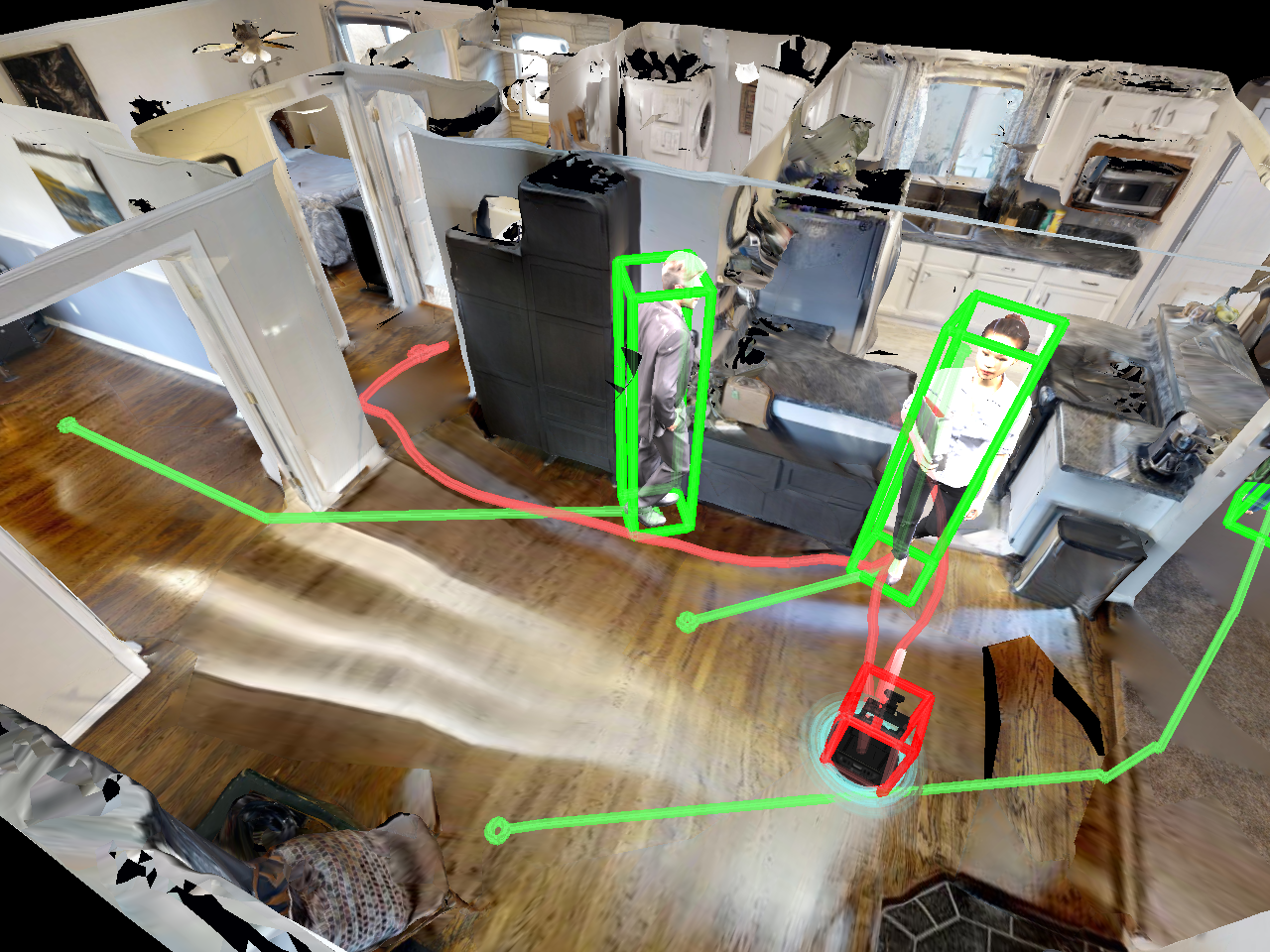}
    \includegraphics[width=.49\linewidth]{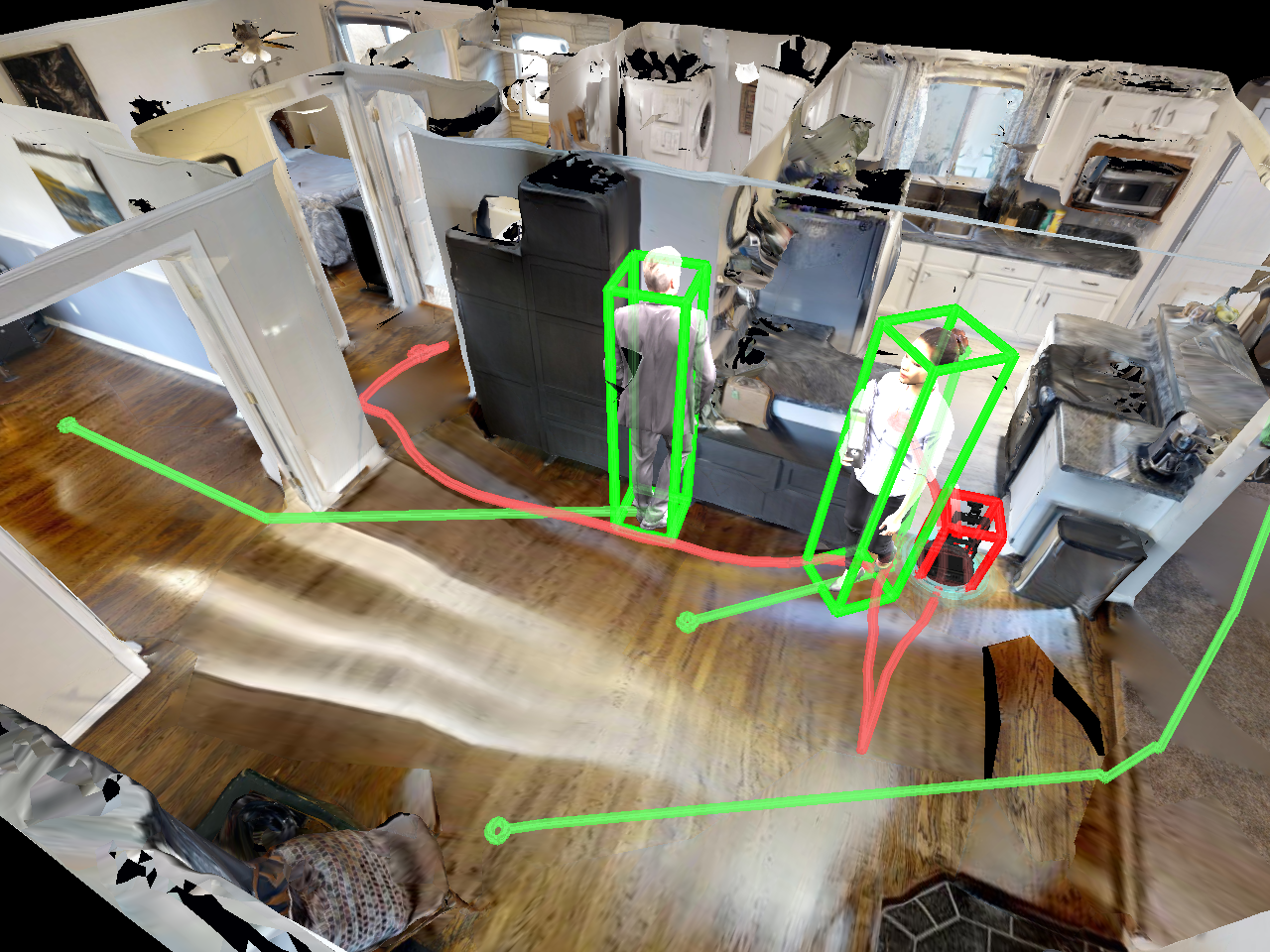}
    \caption{Illustration of an agent-person ``encounter''. From top-left to bottom-right: \emph{i}) episode starts; \emph{ii}) the embodied agent/robot sees a person; \emph{iii}) it moves back to avoid a collision; \emph{iv}) it reaches the goal by avoiding the person.\vspace{-7pt}}
    \label{fig:teaser}
\end{figure}

Navigating safely in a dynamic scenario populated by humans who are moving in the same environment is necessary for embodied agents such as home assistants robots.
To do so, as depicted in Figure \ref{fig:teaser}, the agent should be able to dynamically and interactively navigate the environment by avoiding static objects and moving persons.

Recently, the development of photorealistic 3D simulators \cite{savva2019habitat, shen2021igibson, kolve2017ai2} has provided the tools to train embodied agents and experiment in large-scale indoor environments \cite{chang2017matterport, ramakrishnan2021habitat, deitke2022procthor}.
Thanks to these frameworks, several tasks and challenges have been introduced \cite{anderson2018evaluation, eaisurvey, deitke2022retrospectives}, fostering the development of accompanying techniques to solve these tasks.  In particular, in the PointGoal Navigation task (where an agent is required to reach a specific location in an environment), an agent without any sensor/actuation noise trained for billions of steps can obtain almost perfect performance~\cite{wijmans2019dd}. 
Other approaches~\cite{ye2020auxiliary, partsey2022mapping} obtained impressive results even in the presence of noise. 
Another relevant task is Object Goal Navigation, where an agent is required to find and navigate to a specific object given the object category. This task requires both semantic and navigation capabilities, and lead to development of modular approaches based on semantic-maps \cite{chaplot2020object, campari2020exploiting, ramakrishnan2022poni}, as well as end-to-end trained models using reinforcement learning (RL)~\cite{ye2021auxiliary, ramrakhya2022habitat}.

However, despite encouraging progress on the challenging task, all the previously mentioned tasks frame navigation in a fundamentally static environment.
The dynamic element introduced by sentient, moving human beings in the scene forces us to rethink how the current models are designed. 
A good navigation policy must not be just effective (i.e., able to achieve its goal) and efficient (i.e., able to achieve the objective through a close-to-optimal path) but also safe (reaching the destination without harming others).
This social element is included in the Social Navigation Task (SocialNav)~\cite{xia2020interactive, perez2021robot}, where an agent must tackle PointGoal Navigation in simulated indoor environments. 
To tackle this task, \citet{yokoyama2021learning}  introduced a simple but quite effective model that placed first in the iGibson 2021 SocialNav challenge.  However, the approach does not explicitly encode any social behavior in its navigation policy. 
We believe that a clear encoding of human-agent interactions, as well as social behaviors, are required for safe navigation and interaction with humans. 
By modeling the movement of humans, the agent could prevent collisions or dangerous behaviors and adapt its path to the dynamic environment in which it is navigating. 
We encode these ``signals'' by introducing two \textit{Proximity-Aware Tasks}, referred as \emph{risk} and \emph{proximity compass}. These auxiliary tasks model the present and future danger for the agent's action.

Additionally, we define a fine-grained evaluation protocol for the SocialNav task to better analyse the performances of agents during human-agent interactions. Our evaluation protocol is inspired by a similar attempt~\cite{pirk2022protocol} in robotics, which consisted of collecting statistics about specific types of interaction between humans and a robot (through questionnaires). 
We propose an automated evaluation by identifying and characterizing \textit{encounters} between human and agent.  
To do so, we extract short sub-sequences where an interaction with a human becomes a predominant factor influencing navigation, and we establish a set of rules for classifying each encounter based on the type of human-agent spatial relationship through time.
Finally, we also introduce a dataset of episodes on top of HM3D~\cite{ramakrishnan2021habitat} for Embodied Social Navigation to assess our agents in different environments. 

In summary, the contributions of this work are threefold:
(1) A novel architecture for embodied social navigation which is based on Proximity-Aware tasks; we show the effectiveness of the model on two public datasets.
(2) A new encounter-based evaluation protocol for analysing social navigation models.
(3) A set of episodes for evaluating embodied social navigation based on the HM3D dataset (called HM3D-S).

\section{Related Work}
\mypara{Embodied Navigation.}
In the last few years, there has been growing interest in studying indoor navigation in an embodied setting~\cite{deitke2022retrospectives}. This was mainly possible thanks to large-scale datasets consisting of 3D indoor environments~\cite{chang2017matterport, shen2021igibson, ramakrishnan2021habitat}, and to simulators that allow researchers to simulate navigation inside these 3D environments~\cite{savva2019habitat, shen2021igibson, kolve2017ai2}. To study embodied AI, many tasks~\cite{anderson2018evaluation} were proposed including: PointGoal Navigation~\cite{wijmans2019dd}, ObjectGoal Navigation~\cite{batra2020objectnav}, Embodied Question Answering~\cite{eqa_matterport}, and Vision and Language Navigation (VLN)~\cite{anderson2018vision, krantz2020beyond}. To tackle these problems, where an agent operates in static, single-agent environments, modular approaches were proposed~\cite{campari2022online, chaplot2020neural, chaplot2020object, ramakrishnan2022poni}, exploiting SLAM, Path Planning and exploration strategies, and end-to-end RL-trained policies~\cite{multion,  partsey2022mapping, ye2020auxiliary, campari2020exploiting, ye2021auxiliary}, without exploiting any explicit map. 

In this paper, we focus on the Embodied Social Navigation Task~\cite{perez2021robot}, where the agent has to navigate in environments with moving humans. This task was introduced in the iGibson~\cite{xia2020interactive} SocialNav challenge. SocialNav adds new challenges to visual navigation since modeling of human movements are required to avoid collisions. Modular approaches relying on static maps are harder to adapt in this context since humans are constantly moving and therefore harder to track. 


\mypara{Socially-Aware Navigation.} 
Socially (or human) aware representations and models have been studied by researchers in robotics, computer vision and human social behavior analysis~\cite{survey1}.
Work on collision-free multi-agent navigation~\cite{orca,rvo,cadrl,learned} and navigation in dynamic environments~\cite{mpdyn} has been extended to navigation in the presence of  humans~\cite{guzzi2013human, socialforces, sa-cadrl, social-graph, socialattention}. 
\citet{sa-cadrl} employs collision avoidance algorithms like CADRL~\cite{cadrl} and introduces common-sense social rules to reduce uncertainty while minimizing the risk of incurring in the Freezing Robot Problem~\cite{freeze}.  Other work~\cite{socialforces, socialattention}, attempt to model human-agent interaction using techniques like Spatio-temporal graph~\cite{social-graph}.
These methods have been typically tested on minimalist simulation environments that provide complete knowledge with simple obstacles and often assume collaboration between moving agents. In contrast, we focus on SocialNav in large indoor environments, where the environment is unknown during test time.

\mypara{Egocentric Visual SocialNav.} 
Recently, an increasing number of work has been investigating SocialNav given egocentric visual data. 
Large-scale egocentric datasets of socially compliant demonstrations~\cite{karnan2022socially} and trajectories of agents navigating in social environments~\cite{rudenko2020thor,martin2021jrdb} have enabled the training of social navigation agents via imitation learning.
Despite the usefulness of such large datasets, it's still useful to train agents in simulation, so that it is possible to observe corner cases that not easily collectible in the real world.  Using simulation, agents are typically trained using either imitation learning~\cite{socialim}, end-to-end reinforcement learning~\cite{ yokoyama2021learning}, or a combination of planner with RL~\cite{perez2021robot}.

\citet{socialim} used depth maps to train an agent using imitation learning and studied the policy behavior in a set of simulated interactions. \citet{perez2021robot} focused on constrained indoor environments and uses a combination of a global planner with complete map access and a low-level RL policy that used LIDAR data.
However, this approach requires prior knowledge about the environment for the path planner to work.  In our work, we focus on exploring methods applicable for unknown environments without having to build a map. 
\citet{yokoyama2021learning} proposed an end-to-end RL-trained policy for Social Navigation. This model extracts embeddings from the Depth and the GPS+Compass sensors and feeds them to a GRU, together with the previous action. However, this approach does not model social properties like proximity. Instead, our model learns and uses these concepts with two Proximity-Aware tasks, that help the agent during the episodes. 

\mypara{Evaluation of SocialNav.} Prior work in SocialNav~\cite{gao2022evaluation} are usually evaluated using overall success rate, path efficiency, collision avoidance, or similarity to human paths. Success rate is limited since the agent is dealing with humans and it is preferable to navigate safely in order to avoid collisions, even if this means having lower success rate. In the iGibson challenge, SocialNav is also evaluated using STL (success weighted by time length) and PSC (personal space compliance), the percent of timesteps that the agent is more than a threshold distance (0.5m) from pedestrians. In contrast to these metrics, we propose a fine-grained evaluation based on the types of \emph{encounters} the agent has with pedestrians. 

While there has been attempts at fine-grained evaluation for Social Navigation, there is no standard automatic evaluation based on agent-pedestrian encounters. Most notably, \citet{pirk2022protocol} defined an evaluation protocol for social agents based on human questionnaires. This cannot be easily applied to simulations, where test sets contain thousands of episodes. In this paper, we propose an \emph{automatic} evaluation protocol to measure the social capabilities of our models (see Section \ref{sec:eval_prot}).

\section{Background: the SocialNav Task}
In Embodied Social Navigation~\cite{xia2020interactive, perez2021robot, yokoyama2021learning}, as in PointGoal Navigation, the agent aims to reach a target location, but a collision with a human subject constitutes a failure and will terminate the episode.
An episode $e$ is characterized by the agent trajectory $\boldsymbol{\alpha}$, the tuple of human trajectories $(\mathbf{p}^{i})$, and the target goal $g$. 
The agent trajectory $\boldsymbol{\alpha}$
is a sequence of positions and rotations of the agent from the beginning to the end of the episode $t_\text{end}$. Formally,
$    \boldsymbol{\alpha}=\{\alpha_t\}_{t \in [0, t_\text{end}]}$
where $\alpha_t \in SE(2)$\footnote{SE(2) is the 2-dimensional special euclidean group.} is the 2D translation and rotation of the agent with respect to the origin at time $t$.
Similarly, the trajectories of humans in the same episode are sequences of positions and rotations associated with the $i$-th human.
Formally, 
$\mathbf{p}^{i}=\{p_t^{i}\}_{t \in [0, t_\text{end}]} \; \forall i \in \mathcal{P}$
with $p_t^{i} \in SE(2)$.
In our simulation, the movement of each person is constrained by a start point and an endpoint, with the person moving back and forth between those two points along the shortest path.

The target goal $g \in \mathcal{G}$ is specified by the 2D position in world coordinates.
The agent must at any point in time provide an action $(\text{lin}\_\text{vel}, \text{ang}\_\text{vel}) \in [-1,+1]^2$, representing the normalized linear forward velocity and the normalized clockwise rotational angular velocity (where +1 is the max velocity and -1 the max backward/counter-clockwise velocity).
The stop action is called when the agent is within 0.2 meters from the target goal point. The agent has 500 actions (or steps) to reach the target location. If it collides with a human, the episode terminates immediately.

\section{Method}
\begin{figure*}
    \centering
    \includegraphics[width=0.92\linewidth]{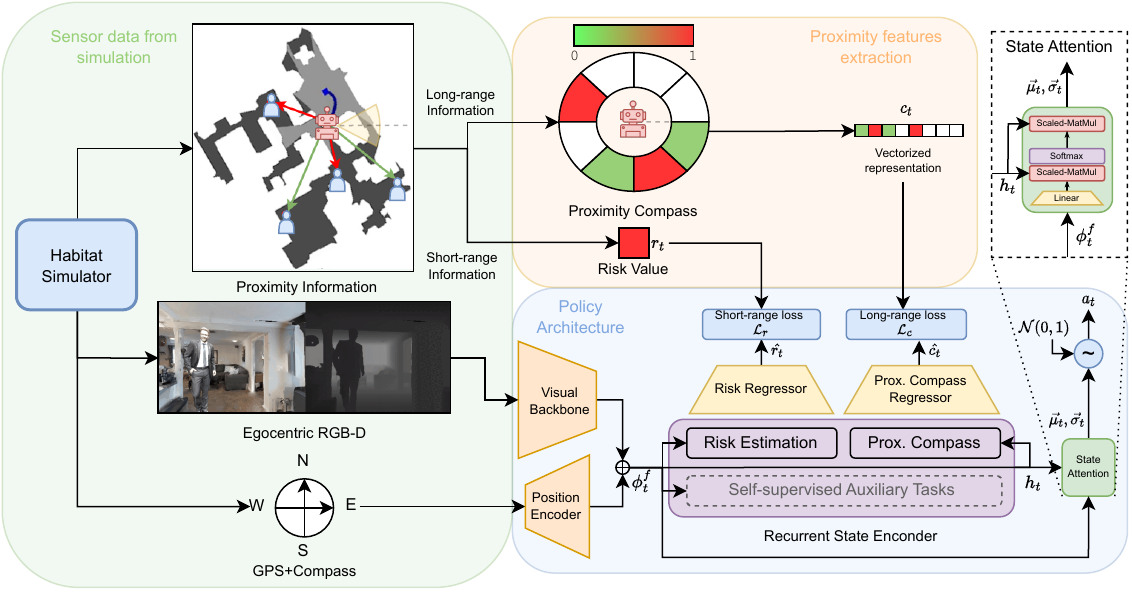}
    \caption{Pipeline and model overview. \textit{Proximity information} is extracted from Habitat Simulator (left rectangle) and is processed through a \textit{Proximity Feature extraction} procedure (top-right). The policy (bottom-right) uses RGB-D and GPS+Compass data as input and, during training, is conditioned by the extracted proximity features.\vspace{-5pt}}
    \label{fig:scheme}
\end{figure*}

\paragraph{Overview.}
Figure \ref{fig:scheme} shows an outline of our framework. It comprises two main modules: (i) \textit{Proximity feature extraction}, and (ii) \textit{Policy architecture}.
The \textit{Proximity feature extraction} module refines proximity information obtained from the simulator to extract features that describe some aspect of social interactions (ground truth proximity features).
The \textit{Policy architecture} extracts from the RGB-D and the GPS+Compass sensors an embedding that serves as input for our Proximity-Aware tasks. These tasks refine this embedding and create \textit{n} task embeddings (one per task) which are then fused together through state attention. An action is sampled from the state attention output. 

\subsection{Policy Architecture}
Our policy network comprises the following modules: \emph{i)} two encoders (the \textit{Visual backbone} and the \textit{Position Encoder}) that create an embedding from the RGB-D and the GPS+Compass sensors; \emph{ii)} a \textit{Recurrent State Encoder} that accumulates such embedding through a series of recurrent units; \emph{iii)} a \textit{State Attention} module that fuses the outputs of such units through an attention mechanism to produce the action the robot has to perform.

We encode each RGB-D frame $x_t$ 
using a CNN (\textit{Visual Backbone}) $f(\cdot)$ to a visual embedding $\phi^v_t = f(x_t)$. 
The position and rotation of the agent $\alpha_t$ are encoded using a linear layer $g(\cdot)$ to obtain the embedding $\phi^p_t = g(\alpha_t)$.  The outputs of these two encoders are then concatenated into the final embedding $\phi^f_t = \phi^v_t \oplus \phi^p_t$.
To accumulate embeddings over time, we follow \citet{ye2020auxiliary}'s design for PointNav and implement our state encoder as a stack of parallel recurrent units. Each unit at each timestep is fed $\phi^f_t$, and outputs its internal state, called \textit{belief}. 

The key idea of having multiple beliefs is that each recurrent unit can focus on a specific navigation aspect. The final decision about what action the robot should take is sampled by weighting each belief according to the situation. For this reason, all beliefs $\mathcal{B}$ are subsequently fused through the \textit{State Attention} module to compute the mean $\vec{\mu_t}$ and standard deviation $\vec{\sigma_t}$ of the normal distribution from which we sample the action $a_t$. Formally, given $\{RU^{(i)}\}_{\forall i \in \mathcal{B}}$ a set of recurrent units, the encoded beliefs $h_t$ are defined as:
\begin{equation}
    \begin{split}
        h_t := \{h_t^{(i)}\}_{\forall i \in \mathcal{B}} \leftarrow  \{RU^{(i)}(h_{t-1}^{(i)}; \phi^f_t)\}_{\forall i \in \mathcal{B}}
    \end{split}
\end{equation}
The fusion mechanism of the state attention module $SA$ is:
\begin{equation*}
    \begin{split}
        \vec{\mu_t}, \vec{\sigma_t} \leftarrow SA(h_t, \phi^f_t) = FC_a(\Attention(h_t, FC_k(\phi^f_t), h_t))
    \end{split}
\end{equation*}
where $\Attention(Q,K,V) \mapsto \Softmax(\frac{QK^T}{\sqrt{d_k}})V$ and $FC_a$ and $FC_k$ are two linear layers.

\subsection{Proximity-Aware Tasks}
With multiple beliefs, we can inject different signals in our embeddings, e.g., social dynamics occurring in an episode. To do so, during training, we condition each belief with a unique auxiliary loss jointly optimized with the action and value ones during the optimization step of the policy network. This is done by processing each belief with a specific type of \textit{Proximity feature}, through a \textit{Regressor network} (see Fig. \ref{fig:socialtask}), that computes our \textit{Proximity-Aware tasks} predictions. 
Each auxiliary task is responsible for predicting the proximity features in the time range $[t,t+k]$, conditioned by the corresponding belief $h_t^{(i)}$ and the sequence of performed actions $\{a_j\}_{j \in [t,t+k]}$, where $k$ is the number of future frames to predict. Formally, for a given sequence of proximity features $\{s_j\}_{j \in [t, t+k]}$, the task aims to optimize the following auxiliary loss:
\begin{equation}
        \mathcal{L_s} = \frac{\sum_{j \in [t,t+k]}\MSE(s_j, \hat{s_j})}{k}\\
        \label{eq:loss}
\end{equation}
where $\{\hat{s_j}\}_{j \in [t,t+k]} = \mathcal{M}(h_t^{(i)}, \{a_j\}_{j \in [t,t+k]})$ and $\mathcal{M}$ is the regressor network.
The proximity features are only fed to the model at training time and regressor networks are detached during evaluation.

We design two types of proximity tasks corresponding to two social features: (i) \textit{Risk Estimation}, and (ii) \textit{Proximity Compass}.
Our design has the benefit of being easily extensible with other, possibly more complex social tasks and to be also compatible with general purpose self-supervised tasks like the ones used in \citet{ye2020auxiliary} (e.g., CPC$|$A \cite{cpca} or ID \cite{id, ye2021auxiliary}).

To exploit different proximity features, we extract from the simulator the relative position of every person w.r.t. the agent. We refer to this data as \textit{Proximity Information}:
$$SI_t \eqdef \{\delta_t^i := (\pos(p^i_t) - \pos(\alpha_t)) \in \mathbb{R}^2\}_{\forall i \in \mathcal{P}}$$
where the function $\pos(\cdot)$ extracts the position from an element of $\boldsymbol{\alpha}$ or $\mathbf{p^i}$. 

\vspace{2pt}
\mypara{Risk Estimation.}
\textit{Risk Estimation} is a Proximity-Aware Task designed to deal with short-range social interactions, to inform the agent about imminent collision dangers. 
Given $SI_t$, we define the \textit{Risk value} as a scalar representing how close the agent and the nearest person are up to a maximum distance $D_r$. This value ranges from 0 (the nearest neighbor is further than $D_r$ meters away) to 1 (the agent and person are colliding). Formally:
\begin{equation}
    \risk_t = \clamp\Big(1 - \frac{\min \{||\delta_t^i||_2 \mid \delta_t^i \in SI_t\}}{D_r}, 0, 1\Big)
\end{equation}
where $\clamp(\cdot, 0, 1)$ limits the value to the $[0,1]$ range.


\vspace{2pt}
\mypara{Proximity Compass.}
The \textit{Proximity-Aware Task} models the long-distance component of social dynamics. This feature captures not only social interaction on a larger area with radius $D_c > D_r$ but also a weak indication of the direction a person may come. Much like humans can make guesses about people's whereabouts based on previous observations, partial knowledge of the environment, and a person's trajectory; we expect our agent to acquire similar knowledge at training time. 

Such information is represented through a \textit{Proximity Compass}. In the compass, north represents the direction the agent is looking, and the quadrant is partitioned into a finite number of non-overlapping sectors. Given each person $i \in \mathcal{P}$, $\theta_{a\rightarrow i}$ represents the angle of the segment connecting the agent to that person w.r.t. the north of the compass. These angles are associated with a specific sector. We compute the risk value for each sector among people in the same quadrant. The entire compass is represented as a vector by unrolling the sequence of sectors from the north going clockwise. Formally if we have $k$ number of equivalent sectors, the vector $\comp_t \in \mathbb{R}^k$ is defined as:
\begin{equation*}
\begin{split}
    \comp_t[j] = \Big[\clamp\Big( 1 -\frac{\min \{||\delta_t^i||_2 \mid \delta_t^i \in \Theta_j\}}{D_c}, 0, 1\Big)\Big]\\
    \text{ with } \; \Theta_j = \Big\{C
    \delta_t^i \in SI_t \mid \theta_{a\rightarrow i} \in \Big[\frac{2\pi}{k}\cdot j , \frac{2\pi}{k}\cdot (j+1)\Big)
    \Big\}
    \\
    \forall j \to[0,k-1]\\
\end{split}
\end{equation*}
\vspace{-15pt}

\subsection{Implementation details}
\begin{figure}
    \centering
    \includegraphics[width=\linewidth]{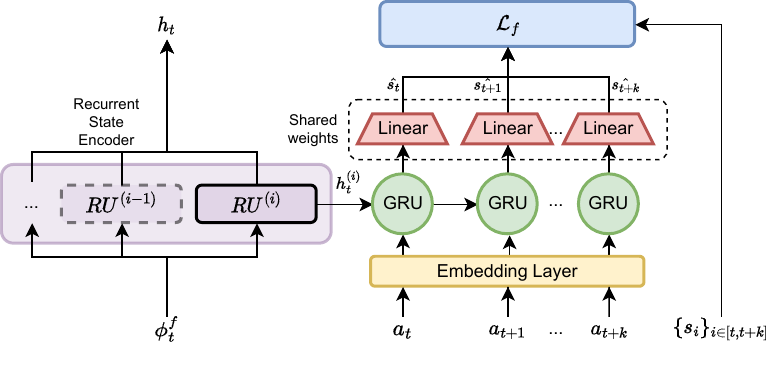}
    \vspace{-24pt}
    \caption{Regressor network. Actions $a_t ... a_{t+k}$ are used as input, and a linear layer processes the GRU's hidden states to obtain the predicted proximity features $\hat{s}_t ... \hat{s}_{t+k}$; $\{s_i\}_{i \in [t,t+k]}$ is the ground truth used by $\mathcal{L}_f$ (from Eq.~\ref{eq:loss}).}
    \label{fig:socialtask}
\end{figure}

The visual backbone used in our experiments is a modified version of a ResNet-18 for RGB-D frames. 
We use the same implementation used in \cite{yokoyama2021learning}; this produces an output embedding of size 512. 
For the agent's position encoder, we use a linear layer with an output size of 32. Our recurrent state encoder is implemented as a stack of single-layer GRUs with a hidden dimension of 512 each. Regressors for both proximity tasks are implemented as single-layer GRUs with hidden size 512. The sequence of actions $\{a_i\}_{i \in [t, t+k]}$ is passed through an embedding layer and then fed to the input of the GRU. The initial hidden state is initialized to $h_t^{(i)}$, and $k$ is set to 4. See Figure \ref{fig:socialtask} for the complete scheme.

Each model was trained on 3 GPUs Nvidia A5000 with 8 parallel environments for each machine using the DD-PPO \cite{wijmans2019dd} algorithm. 
Our reward function is the following:
\begin{equation}
    r_t = -\Delta_d + r_\text{slack} - \beta_{c\land b}\cdot (I_\text{coll} + \cdot I_\text{back}) + \beta_\text{succ}\cdot I_\text{succ}
\end{equation}
where $\Delta_d$ is the potential reward based on the geodesic distance to the goal, $r_\text{slack}$ is the slack reward, and $I_\text{coll}$, $I_\text{back}$, $I_\text{succ}$ represent the indicator functions respectively of a collision with objects in the environment, the linear velocity being less than 0 and success. $\beta_{c\land b}$ and $\beta_\text{succ}$ are coefficients.
We used the same parameters as in \cite{yokoyama2021learning}, that are -0.002 for $r_\text{slack}$, 0.02 for $\beta_{c\land b}$ and 10.0 for $\beta_\text{succ}$.

\section{An evaluation protocol for SocialNav}
\label{sec:eval_prot}

We introduce an evaluation protocol to analyze the performance of navigation agents on the SocialNav task. We group \textit{encounters} between the agent and other pedestrians in the environments into four classes (\textit{frontal}, \textit{intersection}, \textit{blind corner}, \textit{person following}) and propose a set of metrics to analyze the different types of encounters.

\mypara{Encounter.} We formally define an  \textit{encounter} taking place in episode $e$ between the agent and a specific pedestrian $i \in \mathcal{P}$ as a subsequence of trajectories $\boldsymbol{\alpha}$ and $\mathbf{p^{i}}$ in a given timeframe $[t_1,t_2]\subseteq [0,t_\text{end}]$ such that the following constraints are met:
\begin{compactitem}
    \item \textbf{Time Constraint}: the timeframe $[t_1,t_2]$ is larger than a threshold $T_\text{min}$;
    \item \textbf{Spatial Constraint}: the geodesic distance between the agent and person $i$ $\forall t \in [t_1,t_2]$ is less than a threshold $D_\text{max}$;
    \item \textbf{Heading Constraint}: person $i$ is in front of the agent for the first $T_\text{front}$ timesteps. That is, given the agent's heading angle $\theta^a_t$, $\theta^{a \rightarrow i}_t$ the angle of the segment connecting the agent to person $i$ and a threshold $\Theta_\text{max}$, $|\theta^a_t - \theta^{a \rightarrow i}_t| \leq \Theta_\text{max}$ holds $\forall t \in [t_1,t_1+T_\text{front}]$.
\end{compactitem}

\begin{figure}[!t]
\centering
    \begin{subfigure}{.19\textwidth}
        \centering
        \includegraphics[width=\linewidth,trim={0 2cm 0 2cm},clip]{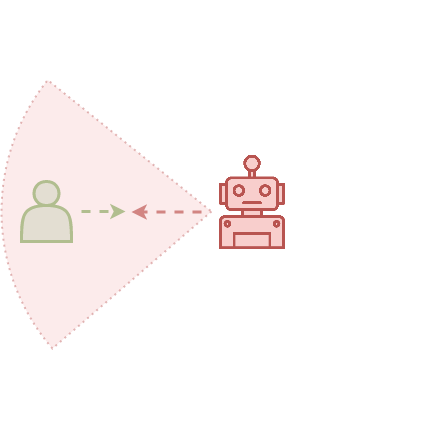}
        \caption{Frontal approach}
    \end{subfigure}
    \begin{subfigure}{.19\textwidth}
        \centering
        \includegraphics[width=\linewidth,trim={0 2cm 0 2cm},clip]{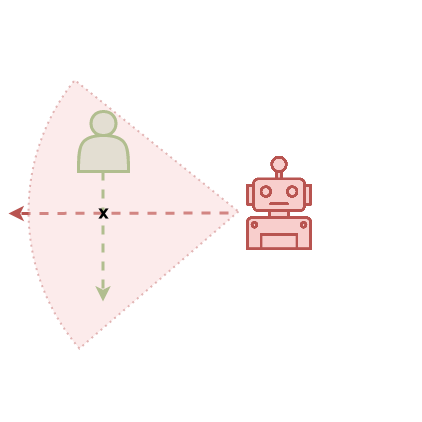}
        \caption{Intersection}
    \end{subfigure}
    \begin{subfigure}{.19\textwidth}
        \centering
        \includegraphics[width=\linewidth]{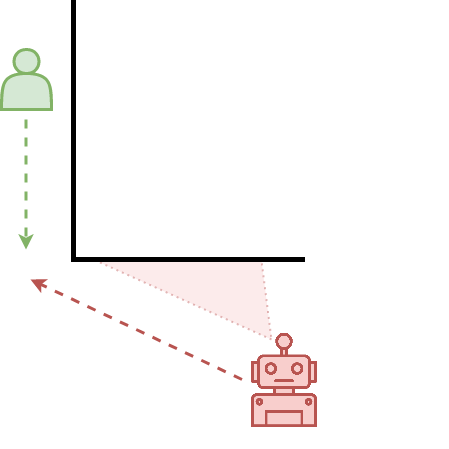}
        \caption{Blind corner}
    \end{subfigure}
    \begin{subfigure}{.19\textwidth}
        \centering
        \includegraphics[width=\linewidth]{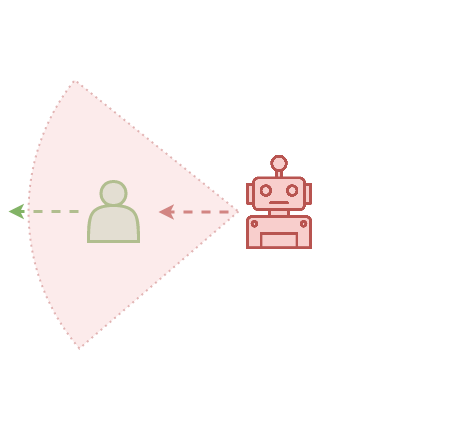}
        \caption{Person following}
    \end{subfigure}
    \caption{The four different classes of encounter. The dashed line represents the general direction of the agent and the person involved. The red area represents the agent's field of view at the beginning of the encounter.\vspace{-5pt}}
    \label{fig:encounters}
\end{figure}


\vspace{2pt}
\mypara{Inclusion rule.}
To characterize encounters between the agent and a human, we devise a heuristic called \textit{inclusion rule (IR)}.  Each IR is defined by the following parameters:
\begin{compactitem}
    \item $\Delta_{t}^i, \Delta_{t}^a$ represent, respectively, an approximation of the general direction of the trajectory of the agent and a person $i$ in the timeframe $[t_1, t]$ with $t \in [t_1, t_2]$, where $t_1, t_2$ are the start and the end timesteps of an encounter;
    \item $\intersect$ is a binary value that is 1 if robot and pedestrian paths intersect and 0 otherwise;
    \item $\blind(t)$ is a time-conditioned binary value indicating whether the agent can see the person at time step $t$;
    \item $\ddiff(t)$: difference between the geodesic and the euclidean distance between the agent and the person.  
\end{compactitem}

\mypara{Encounter classification.} We define four categories of encounters (inspired by \citet{pirk2022protocol}), and their respective \textit{inclusion rule} (see  Figure~\ref{fig:encounters}):
\begin{compactitem}
    \item \textbf{Frontal approach}: The robot and the human come from opposite directions and have trajectories that are roughly parallel. Thus the agent shall deviate slightly to avoid a frontal collision. IR: $(\lnot \blind(t) \; \forall t \in [t_1, t_1 + T_{\view}]) \land \pi-\Delta_{\slack} \leq |\Delta_{t_2}^i - \Delta_{t_2}^a| \leq \pi+\Delta_{\slack}$, where $\Delta_{\slack}$ is a slack value (in radians) on the angle $\pi$ and $T_{\view}$ is the number of initial timesteps in which the person must be visible by the agent.
    \item \textbf{Intersection}: The robot and the human's trajectory intersect at approximately 90°. In this situation, an agent may want to stop and yield to the human or decrease its linear velocity and slightly deviate. IR: $(\lnot \blind(t) \; \forall t \in [t_1, t_1 + T_{\view}]) \land \frac{\pi}{2}-\Delta_{\slack} \leq |\Delta_{t_2}^i - \Delta_{t_2}^a| \leq \frac{\pi}{2}+\Delta_{\slack} \land \intersect$.
    \item \textbf{Blind Corner}: An agent approaches a person from an initially occluded position, like a corner or a narrow doorway. In situations with such a limited visibility, the agent should act cautiously to avoid crashes. IR: $(\blind(t) \; \forall t \in [t_1, t_1 + T_{\blind}]) \land \ddiff(t_1) \leq 0.5$ where $T_{\blind}$ is the number of initial timesteps in which the person must not be visible by the agent.
    \item \textbf{Person following}: A person and the agent travel in the same direction. The agent must maintain a safe distance from the person and a relatively low linear velocity. IR: $(\lnot \blind(t) \; \forall t \in [t_1, t_1 + T_{\blind}]) \land |\Delta_{t_2}^i - \Delta_{t_2}^a| \leq \Delta_{\slack}$.
\end{compactitem}

\mypara{Metrics.}
For each encounter category, we evaluate:
\begin{compactitem}
    \item \textbf{Encounter Survival Rate (ESR)} is the percentage of encounters (in a specific category) without a human collision (e.g., if in the Blind Corner encounter the agent collided with a human in the 20$\%$ of the cases, the ESR will be 80$\%$);
    \item \textbf{Average Linear-Velocity (ALV)} is the average linear velocity of the agent in an encounter; 
    \item \textbf{Average Distance (AD)} is the average distance of the agent w.r.t. the human in an encounter.
\end{compactitem}

\section{Experiments}
\begin{table*}[ht]
    \centering
    \resizebox{\textwidth}{!}{
    \begin{tabular}{|c|c|c|c|c|c|c|c|c|c|c|c|c|c|}
        \hline
         Name & \multicolumn{2}{|c}{Sensors} &  \multicolumn{3}{c}{Aux Tasks} & \multicolumn{2}{c}{Proximity Tasks} & \multicolumn{3}{c}{Metrics (Gibson4+)}  & \multicolumn{3}{c|}{Metrics (HM3D-S)} \\
         \hline
                    & RGB        & Depth                  &  CPCA      & GID         & CPCA/B     & Risk           & Compass           & Success  & SPL   & H-Collisions & Success  & SPL   & H-Collisions \\
         Baseline~\cite{yokoyama2021learning}   &            & \checkmark             &            &             &            &                &            & 72.65\scriptsize{$\pm$1.6}          & 47.43\scriptsize{$\pm$1.2}            &  24.35\scriptsize{$\pm$1.9}            & 62.76\scriptsize{$\pm$2.2} & 36.69\scriptsize{$\pm$1.1} & 29.29\scriptsize{$\pm$2.2}     \\
        Baseline + RGB~\cite{yokoyama2021learning}   &      \checkmark      & \checkmark             &            &             &            &      &         & 74.28\scriptsize{$\pm$1.8}          & 44.84\scriptsize{$\pm$0.7}            &  23.78\scriptsize{$\pm$1.3}            & 61.43\scriptsize{$\pm$0.5} & 34.84\scriptsize{$\pm$0.6}  & 29.23 \scriptsize{$\pm$ 0.7}      \\
         Aux tasks~\cite{ye2020auxiliary}  &\checkmark  & \checkmark             & \checkmark &  \checkmark & \checkmark &                &                 & 73.4\scriptsize{$\pm$2.0}          & 52.08\scriptsize{$\pm$1.4}            &  23.40\scriptsize{$\pm$1.5}            & 63.62\scriptsize{$\pm$1.6} & 42.27\scriptsize{$\pm$1.2} & 24.79\scriptsize{$\pm$2.2}     \\ \hline
         Risk only &\checkmark  & \checkmark             &            &             &            &  \checkmark    &                                         & 74.90\scriptsize{$\pm$1.7}          & 50.25\scriptsize{$\pm$1.1}            &  22.56\scriptsize{$\pm$1.2}            & 66.22\scriptsize{$\pm$1.2}  & 45.26\scriptsize{$\pm$0.8} & 24.47\scriptsize{$\pm$1.7}    \\
         Compass only  &\checkmark  & \checkmark             &            &             &            &      & \checkmark                                    & 75.08\scriptsize{$\pm$1.5}          & 50.55\scriptsize{$\pm$1.0}            &  22.49\scriptsize{$\pm$1.1}            & 67.32\scriptsize{$\pm$1.7} & 45.74\scriptsize{$\pm$1.0} & 23.54\scriptsize{$\pm$1.7}      \\ \hline
         Aux + risk &\checkmark  & \checkmark             & \checkmark & \checkmark  &\checkmark  & \checkmark               &                              & 75.61\scriptsize{$\pm$1.8}          & 51.43\scriptsize{$\pm$0.2}            &  21.04\scriptsize{$\pm$1.4}           & 68.16\scriptsize{$\pm$0.8} & 45.64\scriptsize{$\pm$0.2} & 22.00\scriptsize{$\pm$1.6}    \\
         Aux + compass &\checkmark  & \checkmark             & \checkmark & \checkmark  &\checkmark  &                &\checkmark                             & 75.63\scriptsize{$\pm$1.2}          & 52.60\scriptsize{$\pm$1.6}            &  23.17\scriptsize{$\pm$1.2}           & 67.94\scriptsize{$\pm$1.4} & 45.76\scriptsize{$\pm$1.0} & 23.78\scriptsize{$\pm$2.0}     \\
         Proximity tasks &\checkmark  & \checkmark             &            &             &            &  \checkmark    & \checkmark                           & 76.6\scriptsize{$\pm$1.8}           & 52.81\scriptsize{$\pm$1.2}            &  20.47\scriptsize{$\pm$0.4}   & 68.35\scriptsize{$\pm$0.5} & 45.83\scriptsize{$\pm$0.5} & \textbf{21.72\scriptsize{$\pm$1.2}}    \\   
         Proximity + Aux tasks  &\checkmark  & \checkmark             & \checkmark & \checkmark  &\checkmark  &  \checkmark    &\checkmark                     & \textbf{77.24\scriptsize{$\pm$1.1}}     & \textbf{55.23\scriptsize{$\pm$1.4}}     & \textbf{19.50\scriptsize{$\pm$1.0}}                         & \textbf{70.16\scriptsize{$\pm$1.1}} & \textbf{47.60\scriptsize{$\pm$1.0}} & 22.09\scriptsize{$\pm$1.3}        \\
         \hline
    \end{tabular}
    }
    \caption{Social Navigation evaluation on Gibson4+ and HM3D-S. For each model are listed the type of input data it uses (\textit{Sensors} column) and, eventually, what kind of self-supervised \textit{Aux tasks} or \textit{Proximity tasks} the model employs. The metrics reported are \textit{Success} rate, \textit{SPL} and Human-Collisions Rate (\textit{H-collisions}).\vspace{-7pt}}
    \label{tab:cg_all}
\end{table*}

\paragraph{Datasets and training procedure.}
We performed our experiments using Gibson4+ and HM3D-S, a new dataset based on HM3D and adapted for social navigation~\footnote{Dataset, splits, code, and features will be publicly released.}. 
Gibson4+ contains 86 high-rated scenes taken from the original Gibson dataset \cite{gibson}. For training, we used 64 scenes, while 8 and 14 environments were used for validation and test, respectively. 
HM3D-S is a dataset for Embodied Social Navigation that we generated on top of HM3D~\cite{ramakrishnan2021habitat}. It consists of 900 scenes (800 used for training, 30 for validation, and 70 for test) with, on average, a larger walkable surface compared to Gibson4+. 
We have generated 8M episodes for the training set (10k per scene), 450 for the validation set (15 episodes per environment), and 490 for the test set (7 episodes per environment). Each episode is obtained by selecting a starting point and a goal point for the agent from the navigable area of the environment (such that it exists a navigable path from one to the other). Pedestrians are included as in \cite{yokoyama2021learning}; namely, each person is positioned on a starting point and navigates back and forth to an endpoint with a random linear velocity between $0.45$ and $0.5 m/s$.
On Gibson4+, we trained each model for $\approx$ 100M steps of experience (2.5 days training). On HM3D-S, we fine-tuned our models for $\approx$ 40M steps (1-day training) starting from the final checkpoint obtained on Gibson4+. This was done to reduce the computational cost of each training.

\subsection{Results}
\mypara{Evaluation Metrics.}
We used standard evaluation metrics for Point Goal Navigation such as \textit{Success Rate} and \textit{Success weighted by Path Length} (SPL) (as in previous works~\cite{anderson2018evaluation}). To evaluate its safety properties, we used \textit{Human Collision Rate}, which is the percentage of episodes that end with failure caused by hitting a person. We run all our experiments with five runs to assess the mean and standard deviation for every metric, as done in \cite{yokoyama2021learning}.

\mypara{Baseline models.}
We compared our approach to two baseline models: the model presented in~\cite{yokoyama2021learning} (referred as to \textit{Baseline}) and a ``simplified'' version of our model that only uses a set of 3 self-supervised auxiliary tasks (inspired by \cite{ye2020auxiliary}): namely, 2 CPC$|$A tasks (respectively using 2 and 4 steps) and GID (4 steps). These are referred as to \textit{Aux tasks}. 
Moreover, the \textit{Baseline} model only uses the depth channel as input. Since we believe that RGB input is fundamental for people recognition and trajectory prediction, we also introduce an adapted version of this model that uses all the information from RGB-D frames (called \textit{Baseline+RGB}).

\mypara{Proximity-aware models and auxiliary tasks.}
Firstly, to evaluate each proximity task contribution, we experimented with single-GRU models equipped with just one Proximity-Aware task at a time (i.e., \textit{Risk-only} or \textit{Compass-only}). Since the tasks deal with two different aspects of social navigation (short-range and long-range), we then tried to combine them in a two-GRU model (referred as to \textit{Proximity tasks}). 
Finally, we combined our approaches with the self-supervised auxiliary tasks presented in~\cite{ye2020auxiliary}, which reported state-of-the-art performance on PointGoal Navigation. 
We have thoroughly investigated the benefit of combining them with single proximity tasks (\textit{Aux+Risk} and \textit{Aux+compass}), as well as combining them all in our final model (\textit{Proximity+Aux tasks}).
We now discuss our results and highlight the main takeaways.

\mypara{Performance analysis and comparison to prior work.}
Table \ref{tab:cg_all} reports the social navigation performance (on the test set) for both Gibson4+ and HM3D-S. 
In both cases, \textit{Aux tasks} appears as the strongest of our baselines (highest SPL and lowest Human-Collision for both datasets), reaching comparable performances to single proximity task models while having a higher SPL. 
Our initial hypothesis that integrating the Baseline with an RGB signal could benefit performances was partially supported by the results on Gibson4+. However, the trend shifted on HM3D-S. This happens because of the higher quality of scene reconstruction in Gibson4+ (scenes have been manually rated and are among the best in the original Gibson dataset). Comparatively, HM3D-S has more reconstruction errors that, while leaving depth-only policies unaffected, may impair the performance of RGB-enabled models. 

Moreover, we notice that both models that use just one Proximity-aware task perform similarly on Gibson4+ (sub 0.5\% of difference between metrics). 
However, this changes on HM3D-S, where Compass-only slightly outperforms Risk-only (+1.1\% Success, -0.93\% h-collisions). This difference is expected since the proximity compass task explicitly aims to deal with long-range proximity information. Being HM3D scenes larger in size, the proximity compass role becomes more important.

Adding self-supervised tasks significantly increases SPL and Success performances (both for single-task and all Proximity-tasks models). It also appears to positively affect Human Collision when combined with Risk (-1.52\% in Gibson4+, -2.47\% in HM3D-S). 
We hypothesize that self-supervised tasks, since they are either action-based contrastive tasks (CPC$|$A) or tasks that try to retrieve the inverse dynamics of navigation (GID), help Proximity-Aware models to have smoother trajectories thanks to a more accurate linear and angular velocity dialing. This claim will be substantiated by the fine-grained analysis reported in the next section.
Overall, the best results are obtained by combining all tasks together in the same model.

\subsection{Fine-grained evaluation}
\begin{figure}[t!]
\centering
    \begin{subfigure}{\linewidth}
    \centering
        \includegraphics[width=\linewidth]{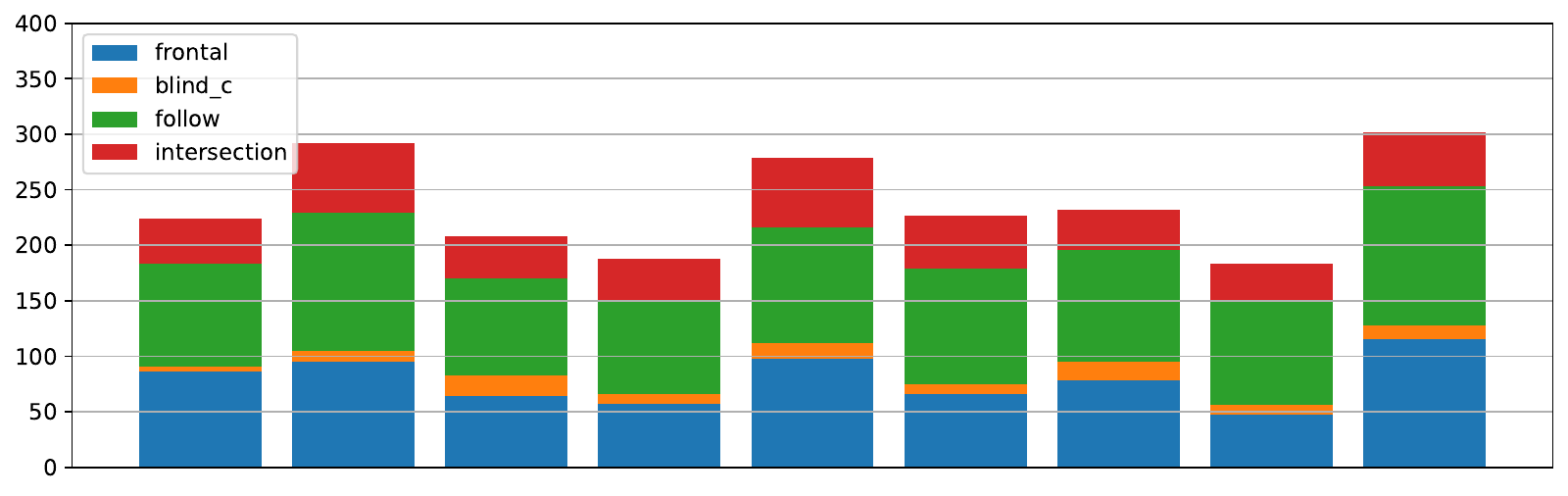}
        \vspace{-18pt}
        \caption{Number of encounters}
    \end{subfigure}
    \begin{subfigure}{\linewidth}
    \centering
        \includegraphics[width=\linewidth]{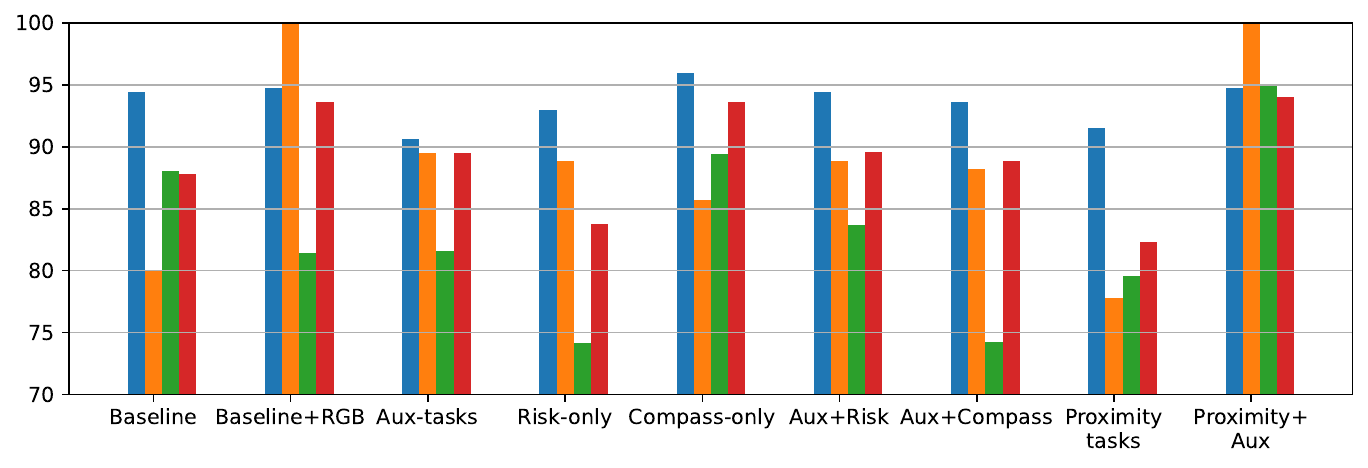}
        \vspace{-18pt}
        \caption{Encounter Survival Rate (ESR)}
    \end{subfigure}
    \caption{Number of encounters and ESR for all the models.\vspace{-7pt}}
    \label{fig:stats}
\end{figure}

\begin{figure}[t!]
\centering
\includegraphics[width=\linewidth,trim={0 0.2cm 0 0.2cm},clip]{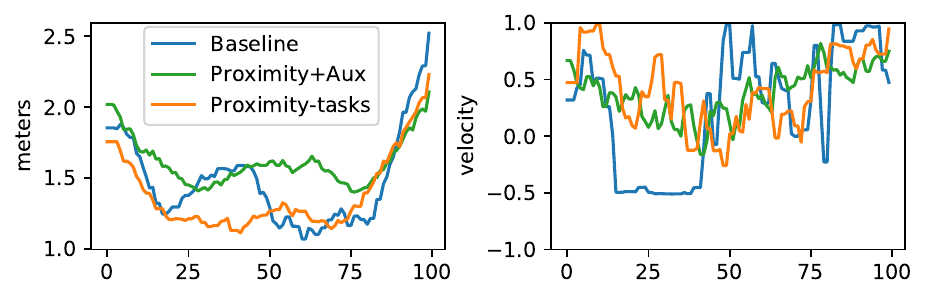}
\caption{Average distance (left) and linear velocity (right) \emph{vs} percentage of completion for Blind-Corner encounters.}
\label{fig:ad_alv}
\end{figure}

We report the results obtained on the Gibson4+ dataset by applying our evaluation protocol (see Section \ref{sec:eval_prot}) to understand better how each model operates and the types of encounters they have. 
Figure \ref{fig:stats} summarizes the statistics, in terms of number of encounters and ESR, collected for each encounter class by different models during 500 randomly sampled validation episodes.

\mypara{Policy behavior analysis.}
Looking at the relationships between the number of encounters and ESR figures, there seem to be two types of policies: a first group with a high number of encounters and a high ESR, and a second group that tends to avoid encounters and has low/medium ESR.
Those two types describe different approaches to social navigation: either risking to interact to access potentially more efficient routes or keeping a safe distance from humans (for example, prioritizing less populated areas, waiting for people to move away before crossing a room).

An example of a policy that avoids encounters is \textit{Risk only}, which has one of the lowest ESR for Following (74.11\%) and Intersection class (86.85\%) and the lowest number of encounters (188 in total). The opposite is true for \textit{Compass only}, which has a high overall ESR for every encounter class and one of the highest numbers of encounters (279 in total).
It is interesting to note how the two best-performing policies, \textit{Proximity tasks} and \textit{Proximity+Aux}, adopt each of these different approaches while remaining comparable in coarse-grained metrics.

\mypara{Reacting to sudden danger.}
A critical ability that a SocialNav policy must possess is the capability to react to immediate and sudden danger in situations with limited visibility. The class of encounters that better represents this is the \textit{blind corner} class. To investigate how our two best models react compared to the baseline, we plotted the average distance (AD) and average linear velocity (ALV) vs percentage of completion of all blind corner encounters (see Figure \ref{fig:ad_alv}). We notice how the ALV curves are not smooth, reflecting uncertainty and high risk. However, while the baseline needs to brake and backtrack ($\approx$ -0.5 ALV between 20\% and 40\% of the episodes), the other models tend to maintain a positive and proportionate ALV velocity throughout the episode. We also see that the \textit{Proximity+Aux} ALV curve is smoother than the one for \textit{Proximity tasks}. The same phenomenon is true for each single Proximity-Aware task model and their self-supervised task counterpart (see supplementary material). This supports the claim that self-supervised tasks provide a smoothing effect on action dynamics under uncertainty.

\mypara{Qualitative results.}
\begin{figure}[t!]
    \centering
    \begin{subfigure}{0.92\linewidth}
        \includegraphics[width=\linewidth]{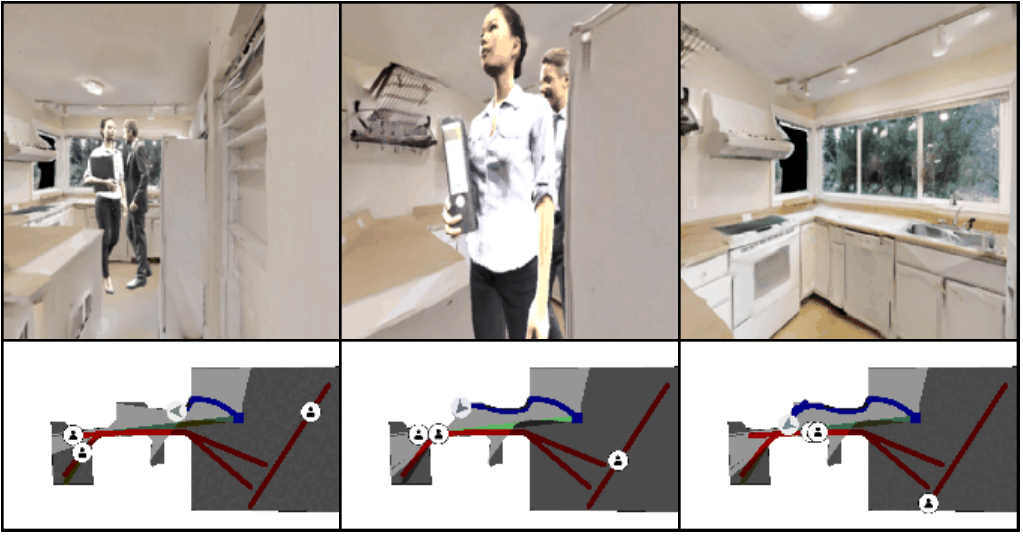}
    \end{subfigure}
    \begin{subfigure}{0.92\linewidth}
        \includegraphics[width=\linewidth]{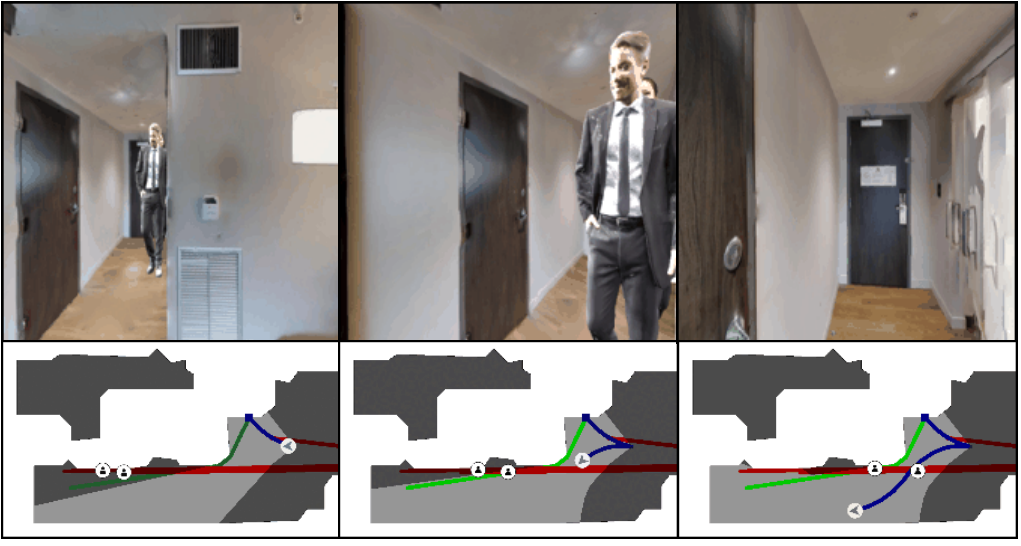}
    \end{subfigure}
    \caption{Two success cases: a frontal encounter (top) and an intersection encounter (bottom).\vspace{-10pt}}
    \label{fig:qual}
\end{figure}
Figure~\ref{fig:qual} shows two qualitative examples of successfully managed encounters in Gibson4+. 
The first example (on top) depicts a Frontal encounter. After seeing people (first frame), the agent moves to the side and yields, letting them move away (second frame). Finally, the agent can reach the goal (third frame). 
In the second example (an intersection encounter), The agent sees a pedestrian (first frame), then it yields, letting the pedestrian pass (second frame), and finally, it continues on its path.

\vspace{-10pt}\paragraph{Limitations.}
A significant limitation of Social navigation in the Embodied-AI literature is the simplicity of trajectory and human movement simulation. To the best of our knowledge, among the ones currently used (such as Habitat, iGibson or AI2Thor), no simulators implement humans with complex navigation patterns and natural-looking animations. This lack implies that social elements like social cues and proxemics are currently not exploited by Embodied-AI models. Our work inherited such problems. However, even under this simplistic approach, the behavior learned by our agents and the properties we discussed in this work represent a strong first step for future work in this area. 

\section{Conclusion}
We introduced a model for Embodied Social Navigation based on two \textit{Proximity-Aware tasks}. 
Our experiments show that exploiting social signals, alone or in combination with self-supervised auxiliary tasks, is an effective strategy in complex and crowded scenarios. 
Our model can avoid the majority of encounters by using only Proximity-Aware tasks. Furthermore, by combining Proximity-Aware and auxiliary tasks~\cite{ye2020auxiliary}, it can
prevent human collisions with high confidence in almost all encounter classes, despite the increased number of them. 
In our future work, we will focus on simulating more natural human behaviors and we will experiment on sim2real domain transfer.

{\small
\bibliographystyle{plainnat}
\setlength{\bibsep}{0pt}
\bibliography{bibliography} 

\begin{thebibliography}{48}
\providecommand{\natexlab}[1]{#1}
\providecommand{\url}[1]{\texttt{#1}}
\expandafter\ifx\csname urlstyle\endcsname\relax
  \providecommand{\doi}[1]{doi: #1}\else
  \providecommand{\doi}{doi: \begingroup \urlstyle{rm}\Url}\fi

\bibitem[Anderson et~al.(2018{\natexlab{a}})Anderson, Chang, Chaplot,
  Dosovitskiy, Gupta, Koltun, Kosecka, Malik, Mottaghi, Savva, and
  Zamir]{anderson2018evaluation}
Peter Anderson, Angel Chang, Devendra~Singh Chaplot, Alexey Dosovitskiy,
  Saurabh Gupta, Vladlen Koltun, Jana Kosecka, Jitendra Malik, Roozbeh
  Mottaghi, Manolis Savva, and Amir~R. Zamir.
\newblock On evaluation of embodied navigation agents.
\newblock \emph{arXiv preprint arXiv:1807.06757}, 2018{\natexlab{a}}.

\bibitem[Anderson et~al.(2018{\natexlab{b}})Anderson, Wu, Teney, Bruce,
  Johnson, S{\"u}nderhauf, Reid, Gould, and van~den Hengel]{anderson2018vision}
Peter Anderson, Qi~Wu, Damien Teney, Jake Bruce, Mark Johnson, Niko
  S{\"u}nderhauf, Ian Reid, Stephen Gould, and Anton van~den Hengel.
\newblock Vision-and-language navigation: Interpreting visually-grounded
  navigation instructions in real environments.
\newblock In \emph{Proc. of the IEEE/CVF Conference on Computer Vision and
  Pattern Recognition (CVPR)}, 2018{\natexlab{b}}.

\bibitem[Aoude et~al.(2013)Aoude, Luders, Joseph, Roy, and How]{mpdyn}
Georges~S. Aoude, Brandon~D. Luders, Joshua~M. Joseph, Nicholas Roy, and
  Jonathan~P. How.
\newblock Probabilistically safe motion planning to avoid dynamic obstacles
  with uncertain motion patterns.
\newblock \emph{Autonomous Robots}, 35\penalty0 (1):\penalty0 51--76, 2013.

\bibitem[Batra et~al.(2020)Batra, Gokaslan, Kembhavi, Maksymets, Mottaghi,
  Savva, Toshev, and Wijmans]{batra2020objectnav}
Dhruv Batra, Aaron Gokaslan, Aniruddha Kembhavi, Oleksandr Maksymets, Roozbeh
  Mottaghi, Manolis Savva, Alexander Toshev, and Erik Wijmans.
\newblock {ObjectNav} revisited: On evaluation of embodied agents navigating to
  objects.
\newblock \emph{arXiv preprint arXiv:2006.13171}, 2020.

\bibitem[Berg et~al.(2011)Berg, Guy, Lin, and Manocha]{orca}
Jur van~den Berg, Stephen~J Guy, Ming Lin, and Dinesh Manocha.
\newblock Reciprocal n-body collision avoidance.
\newblock In \emph{Robotics Research}, pages 3--19. Springer, 2011.

\bibitem[Campari et~al.(2020)Campari, Eccher, Serafini, and
  Ballan]{campari2020exploiting}
Tommaso Campari, Paolo Eccher, Luciano Serafini, and Lamberto Ballan.
\newblock Exploiting scene-specific features for object goal navigation.
\newblock In \emph{Proc. of the European Conference on Computer Vision
  Workshops (ECCVW)}, 2020.

\bibitem[Campari et~al.(2022)Campari, Lamanna, Traverso, Serafini, and
  Ballan]{campari2022online}
Tommaso Campari, Leonardo Lamanna, Paolo Traverso, Luciano Serafini, and
  Lamberto Ballan.
\newblock Online learning of reusable abstract models for object goal
  navigation.
\newblock In \emph{Proc. of the IEEE/CVF Conference on Computer Vision and
  Pattern Recognition (CVPR)}, 2022.

\bibitem[Chang et~al.(2017)Chang, Dai, Funkhouser, Halber, Niessner, Savva,
  Song, Zeng, and Zhang]{chang2017matterport}
Angel Chang, Angela Dai, Thomas Funkhouser, Maciej Halber, Matthias Niessner,
  Manolis Savva, Shuran Song, Andy Zeng, and Yinda Zhang.
\newblock {Matterport3D}: Learning from {RGB-D} data in indoor environments.
\newblock In \emph{Proc. of the International Conference on 3D Vision (3DV)},
  2017.

\bibitem[Chaplot et~al.(2020{\natexlab{a}})Chaplot, Gandhi, Gupta, and
  Salakhutdinov]{chaplot2020object}
Devendra~Singh Chaplot, Dhiraj~Prakashchand Gandhi, Abhinav Gupta, and Ruslan
  Salakhutdinov.
\newblock Object goal navigation using goal-oriented semantic exploration.
\newblock In \emph{Proc. of Advances in Neural Information Processing Systems
  (NeurIPS)}, 2020{\natexlab{a}}.

\bibitem[Chaplot et~al.(2020{\natexlab{b}})Chaplot, Salakhutdinov, Gupta, and
  Gupta]{chaplot2020neural}
Devendra~Singh Chaplot, Ruslan Salakhutdinov, Abhinav Gupta, and Saurabh Gupta.
\newblock Neural topological slam for visual navigation.
\newblock In \emph{Proc. of the IEEE/CVF Conference on Computer Vision and
  Pattern Recognition (CVPR)}, 2020{\natexlab{b}}.

\bibitem[Chen et~al.(2019)Chen, Liu, Kreiss, and Alahi]{socialattention}
Changan Chen, Yuejiang Liu, Sven Kreiss, and Alexandre Alahi.
\newblock Crowd-robot interaction: Crowd-aware robot navigation with
  attention-based deep reinforcement learning.
\newblock In \emph{Proc. of the IEEE International Conference on Robotics and
  Automation (ICRA)}, 2019.

\bibitem[Chen et~al.(2017{\natexlab{a}})Chen, Everett, Liu, and How]{sa-cadrl}
Yu~Fan Chen, Michael Everett, Miao Liu, and Jonathan~P. How.
\newblock Socially aware motion planning with deep reinforcement learning.
\newblock In \emph{Proc. of the IEEE/RSJ International Conference on
  Intelligent Robots and Systems (IROS)}, 2017{\natexlab{a}}.

\bibitem[Chen et~al.(2017{\natexlab{b}})Chen, Liu, Everett, and How]{cadrl}
Yu~Fan Chen, Miao Liu, Michael Everett, and Jonathan~P. How.
\newblock Decentralized non-communicating multiagent collision avoidance with
  deep reinforcement learning.
\newblock In \emph{Proc. of the IEEE International Conference on Robotics and
  Automation (ICRA)}, 2017{\natexlab{b}}.

\bibitem[Deitke et~al.(2022{\natexlab{a}})Deitke, Batra, Bisk, Campari, Chang,
  Chaplot, Chen, D'Arpino, Ehsani, Farhadi, et~al.]{deitke2022retrospectives}
Matt Deitke, Dhruv Batra, Yonatan Bisk, Tommaso Campari, Angel~X. Chang,
  Devendra~Singh Chaplot, Changan Chen, Claudia~P{\'e}rez D'Arpino, Kiana
  Ehsani, Ali Farhadi, et~al.
\newblock Retrospectives on the embodied ai workshop.
\newblock \emph{arXiv preprint arXiv:2210.06849}, 2022{\natexlab{a}}.

\bibitem[Deitke et~al.(2022{\natexlab{b}})Deitke, VanderBilt, Herrasti, Weihs,
  Salvador, Ehsani, Han, Kolve, Farhadi, Kembhavi, et~al.]{deitke2022procthor}
Matt Deitke, Eli VanderBilt, Alvaro Herrasti, Luca Weihs, Jordi Salvador, Kiana
  Ehsani, Winson Han, Eric Kolve, Ali Farhadi, Aniruddha Kembhavi, et~al.
\newblock {Procthor: Large-scale Embodied AI using procedural generation}.
\newblock \emph{arXiv preprint arXiv:2206.06994}, 2022{\natexlab{b}}.

\bibitem[Ferrer et~al.(2013)Ferrer, Garrell, and Sanfeliu]{socialforces}
Gonzalo Ferrer, Anais Garrell, and Alberto Sanfeliu.
\newblock Social-aware robot navigation in urban environments.
\newblock In \emph{Proc. of thr European Conference on Mobile Robots}, 2013.

\bibitem[Gao and Huang(2022)]{gao2022evaluation}
Yuxiang Gao and Chien-Ming Huang.
\newblock Evaluation of socially-aware robot navigation.
\newblock \emph{Frontiers in Robotics and AI}, 8:\penalty0 420, 2022.

\bibitem[Guo et~al.(2018)Guo, Azar, Piot, Pires, and Munos]{cpca}
Zhaohan~Daniel Guo, Mohammad~Gheshlaghi Azar, Bilal Piot, Bernardo~A. Pires,
  and R{\'e}mi Munos.
\newblock Neural predictive belief representations.
\newblock \emph{arXiv preprint arXiv:1811.06407}, 2018.

\bibitem[Guzzi et~al.(2013)Guzzi, Giusti, Gambardella, Theraulaz, and
  Di~Caro]{guzzi2013human}
J{\'e}r{\^o}me Guzzi, Alessandro Giusti, Luca~M Gambardella, Guy Theraulaz, and
  Gianni~A Di~Caro.
\newblock Human-friendly robot navigation in dynamic environments.
\newblock In \emph{IEEE international conference on robotics and automation},
  pages 423--430. IEEE, 2013.

\bibitem[Karnan et~al.(2022)Karnan, Nair, Xiao, Warnell, Pirk, Toshev, Hart,
  Biswas, and Stone]{karnan2022socially}
Haresh Karnan, Anirudh Nair, Xuesu Xiao, Garrett Warnell, S{\"o}ren Pirk,
  Alexander Toshev, Justin Hart, Joydeep Biswas, and Peter Stone.
\newblock Socially compliant navigation dataset ({SCAND}): A large-scale
  dataset of demonstrations for social navigation.
\newblock \emph{IEEE Robotics and Automation Letters}, 7\penalty0 (4):\penalty0
  11807--11814, 2022.

\bibitem[Kolve et~al.(2017)Kolve, Mottaghi, Han, VanderBilt, Weihs, Herrasti,
  Gordon, Zhu, Gupta, and Farhadi]{kolve2017ai2}
Eric Kolve, Roozbeh Mottaghi, Winson Han, Eli VanderBilt, Luca Weihs, Alvaro
  Herrasti, Daniel Gordon, Yuke Zhu, Abhinav Gupta, and Ali Farhadi.
\newblock Ai2-thor: An interactive 3d environment for visual ai.
\newblock \emph{arXiv preprint arXiv:1712.05474}, 2017.

\bibitem[Krantz et~al.(2020)Krantz, Wijmans, Majumdar, Batra, and
  Lee]{krantz2020beyond}
Jacob Krantz, Erik Wijmans, Arjun Majumdar, Dhruv Batra, and Stefan Lee.
\newblock Beyond the nav-graph: Vision-and-language navigation in continuous
  environments.
\newblock In \emph{Proc. of the European Conference on Computer Vision (ECCV)},
  2020.

\bibitem[Long et~al.(2017)Long, Liu, and Pan]{learned}
Pinxin Long, Wenxi Liu, and Jia Pan.
\newblock Deep-learned collision avoidance policy for distributed multiagent
  navigation.
\newblock \emph{IEEE Robotics and Automation Letters}, 2\penalty0 (2):\penalty0
  656--663, 2017.

\bibitem[Lu et~al.(2022)Lu, Ruan, and Huang]{social-graph}
Yazhou Lu, Xiaogang Ruan, and Jing Huang.
\newblock Deep reinforcement learning based on social spatial--temporal graph
  convolution network for crowd navigation.
\newblock \emph{Machines}, 10\penalty0 (8):\penalty0 703, 2022.

\bibitem[Martin-Martin et~al.(2021)Martin-Martin, Patel, Rezatofighi, Shenoi,
  Gwak, Frankel, Sadeghian, and Savarese]{martin2021jrdb}
Roberto Martin-Martin, Mihir Patel, Hamid Rezatofighi, Abhijeet Shenoi,
  JunYoung Gwak, Eric Frankel, Amir Sadeghian, and Silvio Savarese.
\newblock {JRDB}: A dataset and benchmark of egocentric robot visual perception
  of humans in built environments.
\newblock \emph{IEEE transactions on pattern analysis and machine
  intelligence}, 2021.

\bibitem[M{\"o}ller et~al.(2021)M{\"o}ller, Furnari, Battiato, H{\"a}rm{\"a},
  and Farinella]{survey1}
Ronja M{\"o}ller, Antonino Furnari, Sebastiano Battiato, Aki H{\"a}rm{\"a}, and
  Giovanni~Maria Farinella.
\newblock A survey on human-aware robot navigation.
\newblock \emph{Robotics and Autonomous Systems}, 145:\penalty0 103837, 2021.

\bibitem[Partsey et~al.(2022)Partsey, Wijmans, Yokoyama, Dobosevych, Batra, and
  Maksymets]{partsey2022mapping}
Ruslan Partsey, Erik Wijmans, Naoki Yokoyama, Oles Dobosevych, Dhruv Batra, and
  Oleksandr Maksymets.
\newblock Is mapping necessary for realistic pointgoal navigation?
\newblock In \emph{Proc. of the IEEE/CVF Conference on Computer Vision and
  Pattern Recognition (CVPR)}, 2022.

\bibitem[Pathak et~al.(2017)Pathak, Agrawal, Efros, and Darrell]{id}
Deepak Pathak, Pulkit Agrawal, Alexei~A. Efros, and Trevor Darrell.
\newblock Curiosity-driven exploration by self-supervised prediction.
\newblock In \emph{Proc. of the International Conference on Machine Learning
  (ICML)}, 2017.

\bibitem[P{\'e}rez-D’Arpino et~al.(2021)P{\'e}rez-D’Arpino, Liu, Goebel,
  Mart{\'\i}n-Mart{\'\i}n, and Savarese]{perez2021robot}
Claudia P{\'e}rez-D’Arpino, Can Liu, Patrick Goebel, Roberto
  Mart{\'\i}n-Mart{\'\i}n, and Silvio Savarese.
\newblock Robot navigation in constrained pedestrian environments using
  reinforcement learning.
\newblock In \emph{Proc. of the IEEE International Conference on Robotics and
  Automation (ICRA)}, 2021.

\bibitem[Pirk et~al.(2022)Pirk, Lee, Xiao, Takayama, Francis, and
  Toshev]{pirk2022protocol}
S{\"o}ren Pirk, Edward Lee, Xuesu Xiao, Leila Takayama, Anthony Francis, and
  Alexander Toshev.
\newblock A protocol for validating social navigation policies.
\newblock \emph{arXiv preprint arXiv:2204.05443}, 2022.

\bibitem[Ramakrishnan et~al.(2021)Ramakrishnan, Gokaslan, Wijmans, Maksymets,
  Clegg, Turner, Undersander, Galuba, Westbury, Chang,
  et~al.]{ramakrishnan2021habitat}
Santhosh~K. Ramakrishnan, Aaron Gokaslan, Erik Wijmans, Oleksandr Maksymets,
  Alex Clegg, John Turner, Eric Undersander, Wojciech Galuba, Andrew Westbury,
  Angel~X. Chang, et~al.
\newblock {Habitat-matterport 3D dataset (HM3D): 1000 large-scale 3D
  environments for embodied AI}.
\newblock \emph{arXiv preprint arXiv:2109.08238}, 2021.

\bibitem[Ramakrishnan et~al.(2022)Ramakrishnan, Chaplot, Al-Halah, Malik, and
  Grauman]{ramakrishnan2022poni}
Santhosh~K. Ramakrishnan, Devendra~Singh Chaplot, Ziad Al-Halah, Jitendra
  Malik, and Kristen Grauman.
\newblock Poni: Potential functions for objectgoal navigation with
  interaction-free learning.
\newblock In \emph{Proc. of the IEEE/CVF Conference on Computer Vision and
  Pattern Recognition (CVPR)}, 2022.

\bibitem[Ramrakhya et~al.(2022)Ramrakhya, Undersander, Batra, and
  Das]{ramrakhya2022habitat}
Ram Ramrakhya, Eric Undersander, Dhruv Batra, and Abhishek Das.
\newblock Habitat-web: Learning embodied object-search strategies from human
  demonstrations at scale.
\newblock In \emph{Proc. of the IEEE/CVF Conference on Computer Vision and
  Pattern Recognition (CVPR)}, 2022.

\bibitem[Rudenko et~al.(2020)Rudenko, Kucner, Swaminathan, Chadalavada, Arras,
  and Lilienthal]{rudenko2020thor}
Andrey Rudenko, Tomasz~P Kucner, Chittaranjan~S Swaminathan, Ravi~T
  Chadalavada, Kai~O Arras, and Achim~J Lilienthal.
\newblock Th{\"o}r: Human-robot navigation data collection and accurate motion
  trajectories dataset.
\newblock \emph{IEEE Robotics and Automation Letters}, 5\penalty0 (2):\penalty0
  676--682, 2020.

\bibitem[Savva et~al.(2019)Savva, Kadian, Maksymets, Zhao, Wijmans, Jain,
  Straub, Liu, Koltun, Malik, et~al.]{savva2019habitat}
Manolis Savva, Abhishek Kadian, Oleksandr Maksymets, Yili Zhao, Erik Wijmans,
  Bhavana Jain, Julian Straub, Jia Liu, Vladlen Koltun, Jitendra Malik, et~al.
\newblock {Habitat: A platform for embodied AI research}.
\newblock In \emph{Proc. of the IEEE/CVF International Conference on Computer
  Vision (ICCV)}, 2019.

\bibitem[Shen et~al.(2021)Shen, Xia, Li, Mart{\'\i}n-Mart{\'\i}n, Fan, Wang,
  P{\'e}rez-D’Arpino, Buch, Srivastava, Tchapmi, et~al.]{shen2021igibson}
Bokui Shen, Fei Xia, Chengshu Li, Roberto Mart{\'\i}n-Mart{\'\i}n, Linxi Fan,
  Guanzhi Wang, Claudia P{\'e}rez-D’Arpino, Shyamal Buch, Sanjana Srivastava,
  Lyne Tchapmi, et~al.
\newblock {iGibson 1.0: a simulation environment for interactive tasks in large
  realistic scenes}.
\newblock In \emph{Proc. of the IEEE/RSJ International Conference on
  Intelligent Robots and Systems (IROS)}, 2021.

\bibitem[Tai et~al.(2018)Tai, Zhang, Liu, and Burgard]{socialim}
Lei Tai, Jingwei Zhang, Ming Liu, and Wolfram Burgard.
\newblock Socially compliant navigation through raw depth inputs with
  generative adversarial imitation learning.
\newblock In \emph{Proc. of the IEEE International Conference on Robotics and
  Automation (ICRA)}, 2018.

\bibitem[Trautman and Krause(2010)]{freeze}
Peter Trautman and Andreas Krause.
\newblock Unfreezing the robot: Navigation in dense, interacting crowds.
\newblock In \emph{Proc. of the IEEE/RSJ International Conference on
  Intelligent Robots and Systems (IROS)}, 2010.

\bibitem[Van~den Berg et~al.(2008)Van~den Berg, Lin, and Manocha]{rvo}
Jur Van~den Berg, Ming Lin, and Dinesh Manocha.
\newblock Reciprocal velocity obstacles for real-time multi-agent navigation.
\newblock In \emph{Proc. of the IEEE International Conference on Robotics and
  Automation (ICRA)}, 2008.

\bibitem[Wani et~al.(2020)Wani, Patel, Jain, Chang, and Savva]{multion}
Saim Wani, Shivansh Patel, Unnat Jain, Angel~X. Chang, and Manolis Savva.
\newblock {MultiON: Benchmarking semantic map memory using multi-object
  navigation}.
\newblock In \emph{Proc. of Advances in Neural Information Processing Systems
  (NeurIPS)}, 2020.

\bibitem[Wijmans et~al.(2019{\natexlab{a}})Wijmans, Datta, Maksymets, Das,
  Gkioxari, Lee, Essa, Parikh, and Batra]{eqa_matterport}
Erik Wijmans, Samyak Datta, Oleksandr Maksymets, Abhishek Das, Georgia
  Gkioxari, Stefan Lee, Irfan Essa, Devi Parikh, and Dhruv Batra.
\newblock {E}mbodied {Q}uestion {A}nswering in {P}hotorealistic {E}nvironments
  with {P}oint {C}loud {P}erception.
\newblock In \emph{Proc. of the IEEE/CVF Conference on Computer Vision and
  Pattern Recognition (CVPR)}, 2019{\natexlab{a}}.

\bibitem[Wijmans et~al.(2019{\natexlab{b}})Wijmans, Kadian, Morcos, Lee, Essa,
  Parikh, Savva, and Batra]{wijmans2019dd}
Erik Wijmans, Abhishek Kadian, Ari Morcos, Stefan Lee, Irfan Essa, Devi Parikh,
  Manolis Savva, and Dhruv Batra.
\newblock {DD-PPO: Learning near-perfect pointgoal navigators from 2.5 billion
  frames}.
\newblock In \emph{Proc. of the International Conference on Learning
  Representations (ICLR)}, 2019{\natexlab{b}}.

\bibitem[Xia et~al.(2018)Xia, Zamir, He, Sax, Malik, and Savarese]{gibson}
Fei Xia, Amir~R. Zamir, Zhiyang He, Alexander Sax, Jitendra Malik, and Silvio
  Savarese.
\newblock {Gibson Env: Real-World Perception for Embodied Agents}.
\newblock In \emph{Proc. of the IEEE/CVF Conference on Computer Vision and
  Pattern Recognition (CVPR)}, 2018.

\bibitem[Xia et~al.(2020)Xia, Shen, Li, Kasimbeg, Tchapmi, Toshev,
  Mart{\'\i}n-Mart{\'\i}n, and Savarese]{xia2020interactive}
Fei Xia, William~B Shen, Chengshu Li, Priya Kasimbeg, Micael~Edmond Tchapmi,
  Alexander Toshev, Roberto Mart{\'\i}n-Mart{\'\i}n, and Silvio Savarese.
\newblock Interactive gibson benchmark: A benchmark for interactive navigation
  in cluttered environments.
\newblock \emph{IEEE Robotics and Automation Letters}, 5\penalty0 (2):\penalty0
  713--720, 2020.

\bibitem[Ye et~al.(2020)Ye, Batra, Wijmans, and Das]{ye2020auxiliary}
Joel Ye, Dhruv Batra, Erik Wijmans, and Abhishek Das.
\newblock {Auxiliary Tasks Speed Up Learning PointGoal Navigation}.
\newblock In \emph{Proc. of the International Conference on Robot Learning
  (CoRL)}, 2020.

\bibitem[Ye et~al.(2021)Ye, Batra, Das, and Wijmans]{ye2021auxiliary}
Joel Ye, Dhruv Batra, Abhishek Das, and Erik Wijmans.
\newblock {Auxiliary Tasks and Exploration Enable ObjectNav}.
\newblock In \emph{Proc. of the IEEE/CVF International Conference on Computer
  Vision (ICCV)}, 2021.

\bibitem[Yokoyama et~al.(2022)Yokoyama, Luo, Batra, and
  Ha]{yokoyama2021learning}
Naoki Yokoyama, Qian Luo, Dhruv Batra, and Sehoon Ha.
\newblock {Learning Robust Agents for Visual Navigation in Dynamic
  Environments: The Winning Entry of iGibson Challenge 2021}.
\newblock \emph{Proc. of the IEEE/RSJ International Conference on Intelligent
  Robots and Systems (IROS)}, 2022.

\bibitem[Zhu et~al.(2021)Zhu, Zhu, Lee, Liang, and Chang]{eaisurvey}
Fengda Zhu, Yi~Zhu, Vincent Lee, Xiaodan Liang, and Xiaojun Chang.
\newblock Deep learning for embodied vision navigation: A survey.
\newblock \emph{arXiv preprint arXiv:2108.04097}, 2021.

\end{thebibliography}
}


\end{document}